\documentclass[11pt,a4paper]{article}
\usepackage[hyperref]{acl2021}
\usepackage{times}
\usepackage{latexsym}

\usepackage{graphicx}
\usepackage{subfigure}
\usepackage[subfigure]{graphfig}
\usepackage{epstopdf}

\usepackage[utf8]{inputenc}
\usepackage{amsmath}
\usepackage{amsfonts}
\usepackage{multirow}

\usepackage{caption}

\usepackage{microtype}

\title{Ensemble Making Few-Shot Learning Stronger}

\author{Qing Lin, Yongbin Liu*, Wen Wen, Zhihua Tao\\
   \texttt{\{linqiang.fz, yongbinliu03, wenwen.xthn, ttaozhihua\}@gmail.com} \\
        School of Computer, University of South China  \\
  \\}

\date{}

\begin{document}
\maketitle
\begin{abstract}
 Few-shot learning has been proposed and rapidly emerging as a viable means for completing various tasks. Many few-shot models have been widely used for relation learning tasks. However, each of these models has a shortage of capturing a certain aspect of semantic features, for example, CNN on long-range dependencies part, Transformer on local features. It is difficult for a single model to adapt to various relation learning, which results in the high variance problem. Ensemble strategy could be competitive on improving the accuracy of few-shot relation extraction and mitigating high variance risks. This paper explores an ensemble approach to reduce the variance and introduces fine-tuning and feature attention strategies to calibrate relation-level features. Results on several few-shot relation learning tasks show that our model significantly outperforms the previous state-of-the-art models.
\end{abstract}

\section{Introduction}

\begin{table*}
    \centering
    \small
    \begin{tabular}{l|llllllll|l}
    \hline
         Relation & CNN & Incp & GRU & Tran & CNN & Incp & GRU & Trans & Ensemble \\
        ID & Ed & Ed & Ed & Ed & Cosine & Cosine & Cosine & Cosine &  \\ \hline
        P4552 & 86.96 & 88.47 & \textbf{91.29} & 88.80 & 88.56 & 90.21 & 87.57 & 87.22 & \textbf{\textcolor[rgb]{1,0.1,0.1}{91.87}} \\
        P706 & 81.91 & \textbf{82.38} & 81.95 & 79.01 & 79.17 & 75.97 & 77.42 & 78.99 & 82.28 \\
        P176 & 93.90 & 94.33 & 93.40 & 93.32 & 92.84 & \textbf{95.09} & 93.94 & 92.42 & \textbf{\textcolor[rgb]{1,0.1,0.1}{96.12}}\\
        P102 & 96.15 & \textbf{97.02} & 96.72 & 95.45 & 94.17 & 96.32 & 94.54 & 95.89 & \textbf{\textcolor[rgb]{1,0.1,0.1}{97.75}} \\
        P674 & 89.14 & 89.85 & 90.54 & 88.15 & 89.57 & \textbf{91.71} & 90.91 & 87.83 & \textbf{\textcolor[rgb]{1,0.1,0.1}{92.33}} \\
        P101 & 88.30 & \textbf{88.95} & 86.88 & 87.17 & 84.13 & 85.00 & 82.19 & 84.36 & \textbf{\textcolor[rgb]{1,0.1,0.1}{90.17}} \\
        P2094 & 97.69 & 98.16 & 97.99 & 98.96 & 97.87 & 98.34 & 97.03 & \textbf{99.30} & 99.21 \\
        P413 & 96.91 & 96.95 & \textbf{97.75} & 96.76 & 94.74 & 96.68 & 95.22 & 93.05 & \textbf{\textcolor[rgb]{1,0.1,0.1}{98.18}} \\
        P25 & 97.10 & \textbf{97.60} & 96.97 & 96.09 & 96.22 & 96.58 & 95.31 & 96.92 & \textbf{\textcolor[rgb]{1,0.1,0.1}{98.08}} \\
        P921 & \textbf{82.56} & 81.78 & 76.03 & 80.34 & 80.91 & 80.43 & 76.44 & 78.11 & \textbf{\textcolor[rgb]{1,0.1,0.1}{83.54}} \\ \hline
    \end{tabular}
    \caption{\label{table1} Accuracy of each relation of test data in 5-way 5-shot relation classification task using prototypical networks. CNN denotes convolutional neural networks ~\citep{2014Backpropagation} encoder, Incp is inception networks ~\citep{Szegedy2015GoingDW} encoder, GRU is gated recurrent Unit networks ~\citep{2014Learning} encoder, Trans is transformers networks ~\citet{vaswani2017attention06} encoder, Ensemble is our ensemble approach, Ed is Euclidean distance, Cosine is the cosine distance.}
\end{table*}

Few-shot learning can reduce the burden of annotated data and quickly generalize to new tasks without training from scratch. The few-shot learning has become an approach of choice in many natural language processing tasks such as entity recognition and relation classification. There have been many few-shot models proposed in relation extraction task, such as siamese neural network ~\citep{koch2015siamese14}, matching network ~\citep{vinyals2016matching15}, relation network ~\citep{sung2018learning} and prototypical network ~\citep{snell2017prototypical03}. Among these models, the prototypical network is a more efficient and naive approach.

However, representations learning and metrics selecting for relation classification are challenging due to the rich forms of relation expressions in natural language, usually containing local and global, complicated correlations between entities. These are also the leading cause of high variance problem ~\citep{Dhillon2020ABF}. It is problematic that a single model learns the representation for each relation well. This paper aims to learn robust relation representations and similarity metrics from few-shot relation learning.

We propose an ensemble learning approach. We integrate several high accuracy and diversity neural networks to learn the feature representations from each statement's semantics, rather than a sole network. Table~\ref{table1} shows that the four improved prototypical models, which use different neural networks to learn relation and prototype representations respectively, perform on ten relation types where the similarity metric is adopted Euclidean distance and Cosine distance, respectively. We show that the ensemble method performs best on almost all relations, where a collection of diverse representations often serves  better together than a single strong one.

In order to further improve the domain adaptation and leverage the prototype feature sparsity, we explore the fine-tuning and feature attention strategies to calibrate prototypical representations. To adjust to the new relations, we make weight updates using support samples. The fine-tuning method was first proposed in image classification field ~\citep{Dhillon2020ABF,chen2020new,chen2019closerfewshot}. We use the cross-entropy loss function to adjust the weights trained from scratch on an annotated corpus in our fine-tuning strategy. This strategy can significantly improve the accuracy of domain adaptation, especially in cross domain few-shot relation extraction.
In order to better learn the prototypical representation of each relation, we further propose feature attention to alleviate the problem of prototype feature sparsity. The attention mechanism can enhance the classification performance and convergence speed.

We conduct experiments on the FewRel 1.0 dataset~\cite{han2018fewrel08} which comes from wiki(without Domain Adaptation). Then, to show the effect of applying the trained model to other domains, that is, the testing data domain is different from training (with Domain Adaptation); therefore, we choose a new test set PubMed, that comes from a database of biomedical literature and is annotated by FewRel 2.0~\cite{gao2019fewrel09}. Experimental results demonstrate that our ensemble prototypical network significantly outperforms other baseline methods.

\section{Related work}

\begin{figure*}[h]
\centering
\setlength{\abovecaptionskip}{0.cm}
\setlength{\belowcaptionskip}{-0.cm}
\centering
\includegraphics[width=0.8\textwidth]{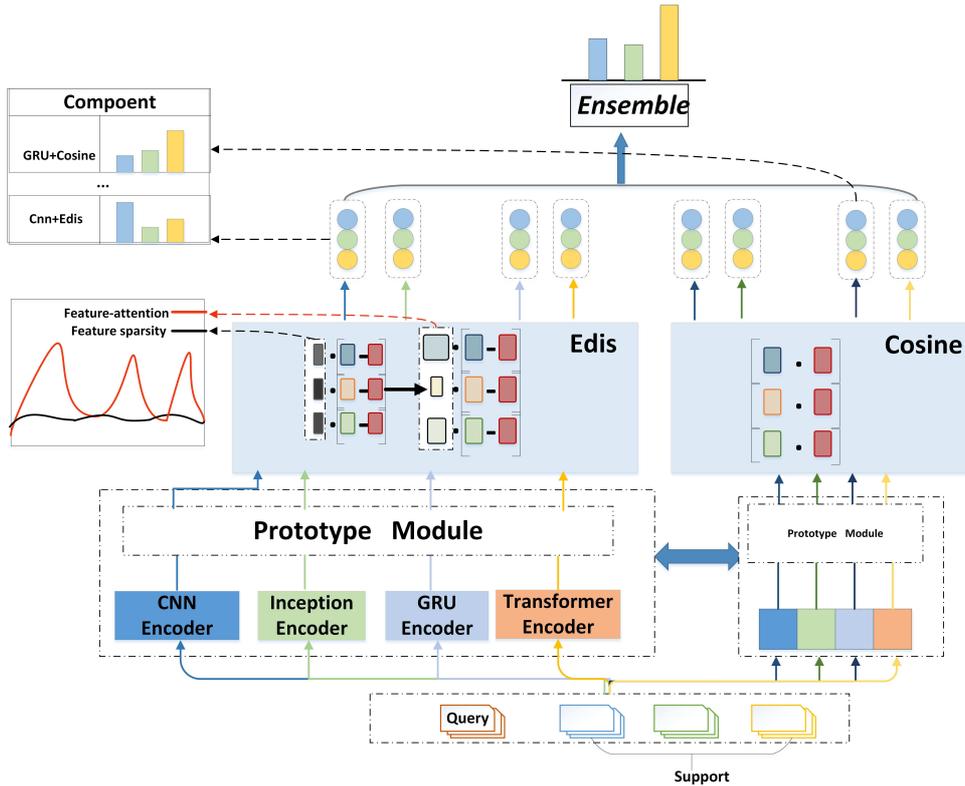}
\caption{Architecture of our proposed ensemble model.}
\label{framework}
\end{figure*}

In this section, we discuss the related work on few-shot learning.

\textbf{Parameters Optimization Learning:} In 2017, a meta network based on gradient optimization has been proposed, which aims to utilize the learned knowledge and then rapidly generalize to new tasks~\citep{munkhdalai2017meta16, ravi2016optimization19}. Meta network entirely relies on a base learner and a meta learner. The base learner gains the meta-information, which includes the information of the input task during dynamic representation learning, and then adaptively updates the parameters for meta learner, while the meta learner memorizes the parameter and acquires the knowledge across different tasks~\cite{vanschoren2018meta17, elsken2020meta}. The agnostic method has been proposed by ~\citet{finn2017model18}. The idea of MAML(Model Agnostic Meta Learning) approach is to learn an initial condition (set of initialization parameter) that is good for fine-tuning on few-shot problems. The few-shot optimization approach ~\citep{ravi2016optimization19} goes further in meta-learning, not only depends on a good initial condition but also utilizes an LSTM based optimizer to help fine-tuning. And then Bayesian Model Agnostic Meta-Learning~\cite{yoon2018bayesian,qu2020few} combines scalable gradient-based meta-learning with nonparametric variational inference in a principled probabilistic framework.

\textbf{Metric Based Few-Shot Learning:} Siamese neural network was applied to few-shot classification by ~\citet{koch2015siamese14}, and it utilized a convolutional architecture to rank the similarity between inputs naturally. Then, matching network ~\citet{vinyals2016matching15} was proposed in 2017. It used some external memories to enhance the neural networks. It added an attention mechanism and a new method named cosine distance as the metric of similarity to predict the relations. MLMAN model ~\cite{ye2019multi} goes further improving the matching network. In 2018, ~\citet{sung2018learning} proposed a relation network for few-shot learning. The relation network learns an embedding and a deep non-linear distance metric for comparing query and sample items. Moreover, the Euclidean distance empirically outperforms the more commonly used cosine similarity on multi-tasks. Thus, a simpler and more efficient model prototypical network was proposed by ~\citet{snell2017prototypical03}. The naive approach used a standard Euclidean distance as the distance function. In 2019, ~\citet{gao2019hybrid02} introduced a hybrid attention-based prototypical network, which is a more efficient prototypical network, and trained a weight matrix for Euclidean distance. These models depended on CNNs, RNNs, and Transformers~\cite{vaswani2017attention06} as the feature extractors. There are always some limitations for a single network to acquiring semantic features.

\textbf{Fine-tuning Methods:} The idea of fine-tuning is used in few-shot learning, which refers to the pre-training model ~\citep{devlin2018bert, Radford2018ImprovingLU, Peters2018DeepCW}. The fine-tuning deep network is a strong baseline for few-shot learning ~\citep{chen2020new,chen2019closerfewshot}. These works connect the softmax cross-entropy loss with cosine distance. In 2020, ~\citet{Dhillon2020ABF} introduced a transductive fine-tuning baseline for few-shot learning. Most of these works have been developed in the image domain but are not widely used in the relation extraction domain. Different from images, the text is more diverse and complicated. So we demand to apply them to few-shot relation learning tasks. 

In the paper, we discuss several factors that affect the robustness of few-shot relation learning. We propose the novel ensemble few-shot learning model that integrates four networks and two metrics into a prototypical network. Further more, our proposed model adopts the fine-tuning to improve the domain adaptation and feature attention strategy to address the problem of feature sparsity.

\section{Our Approach}

In this section, we give a detailed introduction to the implementation of our ensemble few-shot model, as shown in Figure \ref{framework}.

\subsection{Notations and Definitions}
We follow ~\citet{gao2019hybrid02} and ~\citet{snell2017prototypical03} to define our few-shot setting. Few-shot Relation classification(RC) is defined as a task to predict the relation $r$ between the entity pair ($h$, $t$) mentioned in a query instance
$x$. Given a relation $r \in R$ and a small support set of $N$ labeled examples $S$ $=\{(x_{1},y_{1}),...,(x_{N},y_{N})\}$, each $x_{i}\in \mathbb{R}^{D}$  is the $D$-dimensional feature vector
of an example and $y_{i}\in\{1,...,K\}$ is the corresponding label.

The  prototypical network ~\citep{snell2017prototypical03} assumes that there exists a prototype where points cluster around  for  each  class,  and  the  query  point  is  classified  by  using  the  distance function  to  calculate  the  nearest  class  prototype,  which  is  determined  by  the instances in its support set. Given an instance $x$ = $\{e_{1}, . . . ,e_{n}\}$ mentioning two entities, we encode the instance into a low-dimensional embedding $\mathbf{x}=f_{\theta}(x_{i})$   through an embedding function with learnable parameters $\theta$, which are different for different neural networks encode layer. In our ensemble model, we adopt four classical neural networks to learn the  $\mathbf{x}$ respectively. The main idea of prototypical networks is to compute a class representation $\mathbf{c}_{k}$   named prototype.

\begin{equation}
\mathbf{c}_{k}=\frac{1}{\mid S_{k}\mid}\sum\limits_{(x_{i},y_{i})\in S_{k}}\mathbf{x}_{i}
\end{equation}

Given a test point $x$, We can compute a distribution over classes as follow,

\begin{equation}
p_{\theta}(y=k\mid x)=\frac{exp(-d(\mathbf{x},\mathbf{c}_{k}))}{\sum_{k'}exp(-d(\mathbf{x},\mathbf{c}_{k'}))}
\end{equation}

where $d$ is the distance function, which can be either Euclidean distance or Cosine distance. In our paper, both metrics are taken into account in our model. Also, we adopt feature-level attention to compute the $d$, which can achieve better results and convergence speed. 

\subsection{Feature Attention}

The original model uses simple Euclidean distance as the
distance function. In fact, some dimensions are more discriminative for classifying relations in the feature space ~\citep{gao2019hybrid02}. We improve the feature-level attention in the ~\citet{gao2019hybrid02}, and propose the feature attention based on vector subtraction that can find those dimensions more discriminated.

\begin{equation}
a_{j}=\sum_{i=1}^{n} \sum_{l=1,l\neq i}^{n}|x^{i}-x^{l}| /n^2\label{feature attention1}
\end{equation}

where $n$ is the number of sample in support set.

\begin{equation}
w_{j}=-\log \left(a_{j}+\xi \right)\label{feature attention2}
\end{equation}

where $\xi$ is a hyperparameter.

\begin{equation}
d_{ed}\left(x_{q}, \mathbf{c}_{k}\right)=w_{k}\left(x_{q}-\mathbf{c}_{k}\right)^2
\end{equation}

\begin{equation}
d_{cos}\left(x_{q}, \mathbf{c}_{k}\right)=w_{k}\frac{x_{q}^{T} \mathbf{c_{k}}}{|| x_{q}||_{2} \cdot|| \mathbf{c}_{k}||_{2}}
\end{equation}

where $w_{k}$ is the score vector for relation $r_{\mathbf{c}_{k}}$ computed via (\ref{feature attention1}) and (\ref{feature attention2}). The $d_{ed}$ refers to Euclidean distance and the $d_{cos}$ refers to Cosine distance.

\subsection{Learning Ensemble of Deep Networks}

To reduce the high variance of few-shot learning, we use ensemble methods acquiring semantic feature, as in Figure \ref{framework}. Now we discuss the objective functions for learning ensemble prototypes model. During meta-training, each network needs to minimize the cross-entropy loss function over a training dataset:

\begin{equation}
L(\theta)=\frac{1}{n} \sum_{i=1}^{n} \ell\left(y_{i}, f_{\theta}\left(x_{i}\right)\right)+\lambda\|\theta\|_{2}^{2}\label{loss}
\end{equation}

where $f_{\theta}$ is a basic neural networks, which could be CNN, Inceptions, GRU and Transformer in our experiments. The cost function $\ell$ is the cross-entropy between label $y_{i}$ and $f_{\theta}(x_{i})$, and the $\lambda$ is a weight decay parameter.

In our ensemble model, $E$ is the number of ensemble networks. When training each networks $f_{\theta_{e}}$ independently, $e \in E$, each $f_{\theta_{e}}$  would  performance on (\ref{loss}) separately. Due to choosing basic networks are very different, each network learns the semantic feature will typically differ. This is the reason of ensemble model appealing in our paper. 

In our ensemble model, we propose a joint loss function to ensemble each network:

\begin{equation}
\begin{aligned}
L(\tilde{\theta})=& \sum_{e=1}^{E}\left(\frac{1}{n} \sum_{i=1}^{n} \ell\left(y_{i}, f_{\theta_{e}}\left(x_{i}\right)\right)+\lambda\left\|\theta_{j}\right\|_{2}^{2}\right) \\
&+\sum_{e=1}^{E}\left(\frac{1}{n} \sum_{i=1}^{n} \mathbb{H}\left(f_{\theta_{e}}\left(x_{i}\right)\right)\right)
\end{aligned}
\end{equation}

where $\mathbb{H}$ is the Entropy on $f_{\theta_{e}}$, we aim to seek outputs with a peaked posterior. The loss function would help to solve the collaboration of the ensemble networks during training.

\subsection{Fine-tuning Learning}

For our model to recognize novel relation classes, especially for the cross domain, we adopt a fine-tuning strategy to improve our model domain adaptation. 
We use support set $S=\{\{x_{i},y_{i}\}_{i=1}^{n_{1}},...,\{x_{i},y_{i}\}_{i=1}^{n_{K}}\}$ to fine-tuning our model parameters, where ${n_{k}}$ is the number of available samples of the class $k$. we encode $x_{i}$ of the $S$ to the representation and feed to a feed-forward classifier (Softmax layer):

\begin{equation}
p_{i}=Softmax\left(\mathbf{M} \cdot f_{\tilde{\theta}}\left(x_{i}\right)+\mathbf{b}\right)
\end{equation}

where $\mathbf{M}$ and $\mathbf{b}$ denote the learning parameters in the feed-forward layer, and we train it with the cross-entropy loss:

\begin{equation}
\min\sum_{i}CrossEntropy\left(y_{i}, p_{i}\right)\label{opt}
\end{equation}

by (\ref{opt}), we could optimize the parameters of our model. 
By carefully initializing appropriately the parameter $\mathbf{M}$, it is possible to achieve desirable properties of the ensemble ~\citep{Dhillon2020ABF}. So we use each class prototype in the $S$ to initialize $\mathbf{M}$, setting $\mathbf{b}=0$.

\section{Experiments}

In this section, we present the experimental and implementation details of our methods. First, we compare our proposed ensemble model with existing state-of-the-art models on different levels to show the advantages. Secondly, we further study the effect of different parts that are integrated into our ensemble model. Our ensemble model integrates CNN,Inception, GRU and Transformers networks adopting Euclidean and Cosine distance respectively, as shown in Figure \ref{framework}.

\subsection{Datasets}

For various N-way K-shot tasks, we evaluate our proposed our model on two open  benchmarks: FewRel 1.0 ~\citep{han2018fewrel08} and FewRel2.0~\citep{gao-etal-2019-fewrel}, which is shown in Table \ref{datasets}.

\begin{table}[h]
    \centering
    \scalebox{0.6}{
    \begin{tabular}{l|l|l|l|l}
    \hline
        Source Dataset & Source & Apply & Relation \# & Instance \# \\ \hline
        FewRel 1.0& Wiki & Training & 60 & 42,000 \\ 
        ~\citep{han2018fewrel08} & Wiki & Validation & 10 & 7,000 \\
         & Wiki & Testing & 10 & 7,000 \\ \hline
        FewRel 2.0 & Wiki & Training & 64 & 44,800 \\ 
        ~\citep{gao-etal-2019-fewrel} & SemEval-2010 task 8 & Validation & 17 & 8,851 \\
         & PubMed & Testing & 10 & 2,500 \\ \hline
    \end{tabular}}
    \caption{\label{datasets}Datasets}
\end{table}

The FewRel 1.0 dataset has 80 relations and 700 examples for each. Due to the origin test set is hidden, we split the datasets into the training, validation, and test set. To satisfy our experiment setting, we randomly choose 60 relations for the train set, 10 relations for the validation, and the rest 10 relations for testing, and the examples of our test set are disjointed with the validation. However, the train set, validation set, and test set are all come from the Wikipedia corpus; that is, they are in the same domain, which is not practicable enough. Thus, we utilize the data sets in FewRel 2.0, which is cross domain. The train set is from Wiki, which is the same as FewRel 1.0, and it has 64 relations, while the validation set is from SemEval-2010 task 8 with 17 relations, which is annotated on news corpus, and then test on PubMed, which is from the database of biomedical domains and has 10 relations. Moreover, we use accuracy as the evaluation criteria.

\subsection{Experimental Setup}

Our model hyper-parameters are shown in table \ref{parameter}. We randomly select 20 samples from the data set each time as the query for training and use the GloVe 50-Dimensional word vectors as our initial word embeddings. Our model uses the SGD with the weight decay of $10^{-5}$ as optimizer during training and fine-tuning. We perform fine-tuning for 60 epochs without any regularization and updates the weight by the cross-entropy using support samples.

\begin{table}[h]
    \centering
    \scalebox{0.7}{
    \begin{tabular}{l|l}
    \hline
        Batch Size & 4 \\ 
        Query Size & 20 \\ 
        Training Iterations & 30000 \\ 
        Val Step & 2000 \\ 
        Learning Rate & 0.1 \\ 
        Weight Decay & 10-5 \\ 
        Optimizer & SGD \\ 
        $\xi$ & 0.1 \\ 
        Fine-tune iterations & 60 \\ \hline
    \end{tabular}}
    \caption{\label{parameter}Hyper-parameters Setting}
\end{table}

\subsection{Baselines}

Siamese network ~\citep{koch2015siamese14} maps two samples into the same vector space by using two subnetworks respectively and then calculate the distance between two samples by some distance function. FSL Graph Neural Networks (GNN) ~\citep{garcia2017few} maps all support and query into the vertex in the graph and utilizes the graph neural networks to classify. Prototypical network (Proto) ~\citep{snell2017prototypical03} can be classified by measuring the distance between one and all prototypes and selecting the label of the nearest prototype. Snail ~\citep{snail2017A} is a meta learning model that formalizes meta-learning as a sequence-to-sequence problem, using a combination of temporal convolution (TC) and attention mechanism. A hybrid attention-base prototypical network (Proto\_hatt) ~\citep{gao2019hybrid02} is a variant of the prototypical network, which consist of an instance-level attention module and feature-attention module. These are current state-of-the-art few-shot models.

\begin{table*}[h]
    \centering
    \scalebox{0.8}{
    \begin{tabular}{l|lll|lll}
    \hline
        Cross-domain  &  & 5 way  &  &  & 10 way  &  \\ 
       (FewRel2.0) & 1 shot & 5 shot & 10 shot &  1 shot &  5 shot &  10shot \\ \hline
        Siamese  & 39.66 & 47.72 & 53.08 & 27.47 & 33.58 & 38.84 \\ 
        GNN & 35.95 & 46.57 & 52.20 & 22.73 & 29.74 & - \\ 
        Proto & 40.16 & 52.62 & 58.69 & 28.39 & 39.38 & 44.98 \\ 
        Proto\_hatt & 40.78 & 56.81 & 63.72 & 29.26 & 43.18 & 50.36 \\ 
        Bert-pair & \textbf{56.25*} & 67.44* & - & \textbf{43.64*} & 53.17* & - \\ \hline
        Proto\_atten & 41.55 & 55.87 & 62.28 & 29.68 & 42.34 & 48.63 \\ 
        Ensemble\_cosine & 42.74 & 59.12 & 65.89 & 30.93 & 45.51 & 52.39 \\ 
        Ensemble\_edis & 44.04 & 61.41 & 68.07 & 32.13 & 47.88 & 54.70 \\ 
        Ensemble & 44.42 & 61.76 & 68.49 & 32.44 & 48.26 & 55.23 \\ 
        Ensemble\_fine-tuning & / & \textbf{68.32} & \textbf{74.96} & / & \textbf{63.98} & \textbf{70.61} \\ \hline
    \end{tabular}}
    \caption{\label{cross}Results on cross-domain (FewRel2.0).  *Results reported by ~\citet{gao-etal-2019-fewrel},	/ one shot tasks is not suitable for our fine-tuning method.}
\end{table*}

\begin{table*}[h]
    \centering
    \scalebox{0.85}{
    \begin{tabular}{l|lll|lll}
    \hline
        In-domain  &  & 5 way  &  &  & 10 way  &  \\ 
       (FewRel1.0) & 1 shot & 5 shot & 10 shot &  1 shot &  5 shot &  10shot \\ \hline
        Siamese  & 75.76 & 85.80 & 89.04 & 64.58 & 77.42 & 80.30 \\ 
        GNN & 71.18 & 85.71 & 89.25 & 56.01 & 74.33 & - \\ 
        Snail & 72.69 & 84.22 & 85.23 & 58.15 & 68.36 & 73.36 \\         
        Proto\_hatt & 75.45 & 89.97 & 92.03 & 62.64 & 82.29 & 85.74 \\ 
        Proto & 74.01 & 89.46 & 91.55 & 61.30 & 81.66 & 84.87 \\ \hline
        Ensemble\_edis & 79.70 & 92.55 & 94.10 & 69.19 & 86.61 & 89.02 \\ Ensemble\_cosine & \textbf{81.40} & 92.56 & 93.98 & 71.22 & 86.72 & 88.90 \\ 
        Ensemble & 81.35 & \textbf{92.90} & \textbf{94.32} & \textbf{71.29} & \textbf{87.30} & \textbf{89.46} \\ \hline
    \end{tabular}}
    \caption{\label{in}Results on in-domain(FewRel 1.0). - For our GPU: out of memory}
\end{table*}

\subsection{Experimental Results and Discussion}

In this part, we present the comparison results between our proposed model and the typical models of few-shot learning under the same hyper-parameters, which are given in Table \ref{parameter}. 

First, we show the results on FewRel 2.0 which is the cross domain in Table \ref{cross}. As we can see, our ensemble model  has over 3\% improvement for different scenarios, and the maximum number even reach 20\%, when compared with the state-of-the-art model without pre-trained model.

\begin{figure}[h]
\centering
\setlength{\abovecaptionskip}{0.cm}
\setlength{\belowcaptionskip}{-0.cm}
\centering
\includegraphics[width=0.5\textwidth]{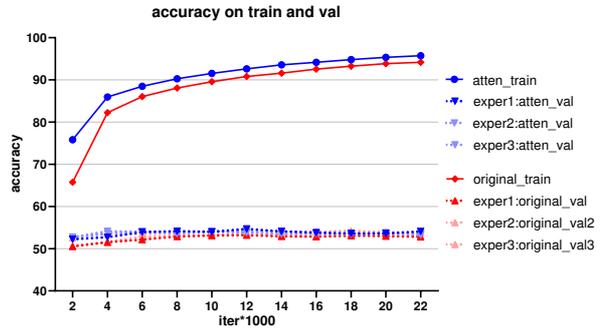}
\caption{Comparsion between proto\_atten and proto.}
\label{zhexian}
\end{figure}

In this experiment, we utilize a feature attention mechanism to focus our ensemble model on the relation-level features. Gave high weights to these dimensions are able to highlight the commonality of the examples in the same relation, which helps to discriminate the similarity and the comparison between the prototype networks with feature attention (Proto\_atten) and the prototype networks without feature attention (Proto) on FewRel 2.0 datasets, the results are shown as Figure \ref{zhexian}. In addition, the Proto\_atten model is a stronger classifier without increasing the number of parameters, which is able to learn feature with more semantics, and then accelerate convergence. 

\begin{figure*}[h]
\begin{minipage}[t]{0.5\textwidth}
\centering
\includegraphics[width=0.6\textwidth]{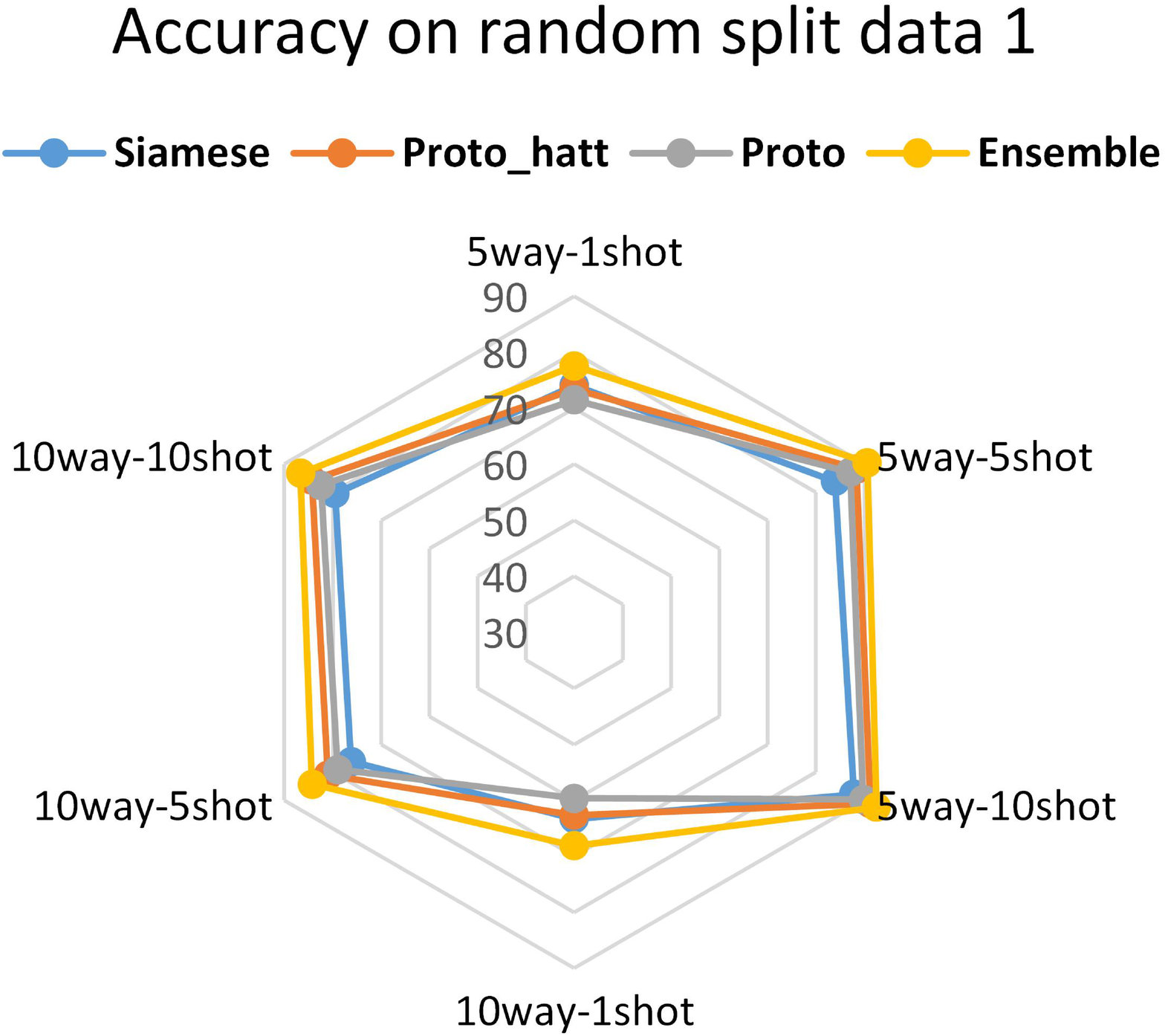}
\end{minipage}
\begin{minipage}[t]{0.5\textwidth}
\centering
\includegraphics[width=0.6\textwidth]{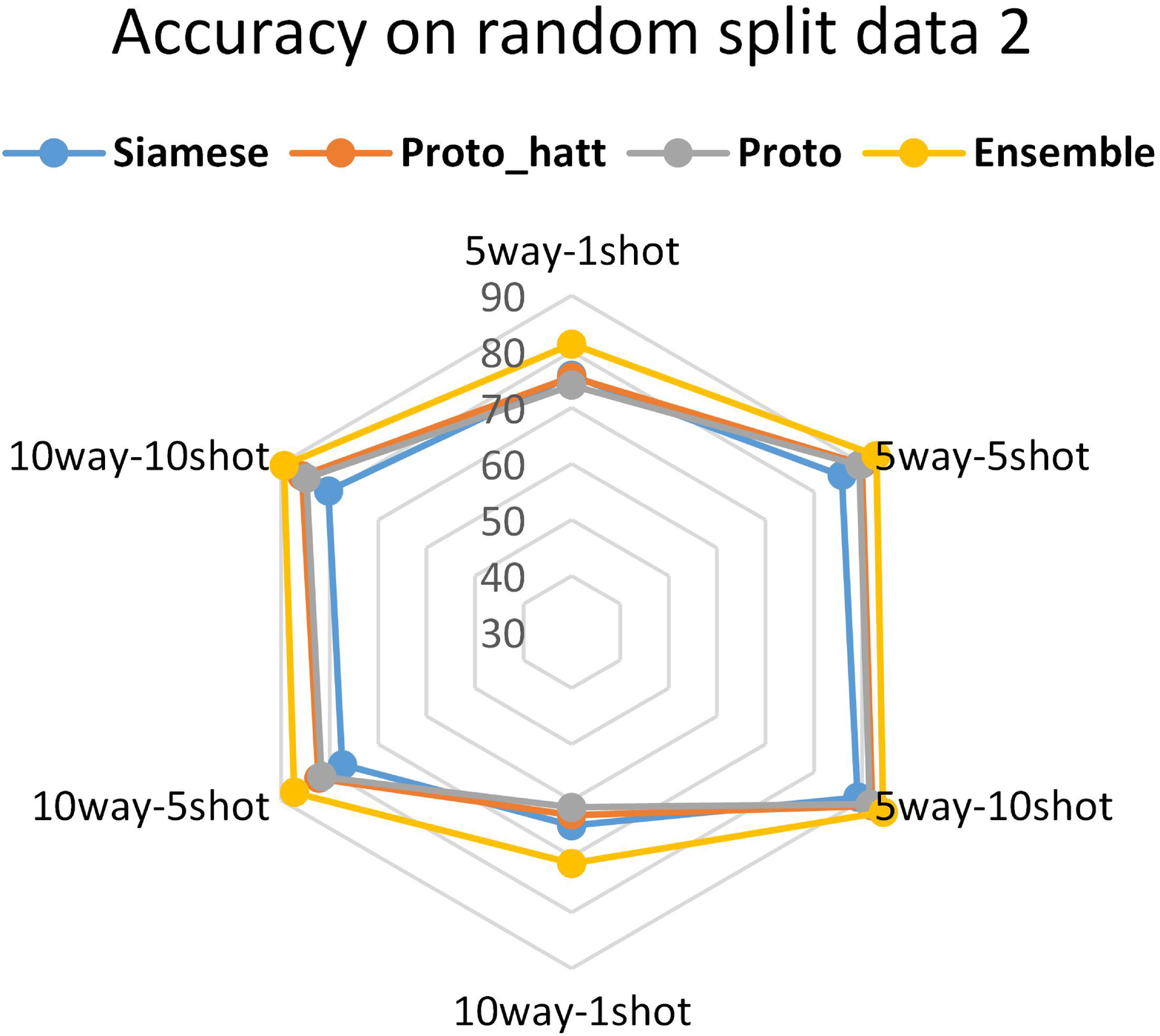}
\end{minipage}

\begin{minipage}[t]{0.5\textwidth}
\centering
\includegraphics[width=0.6\textwidth]{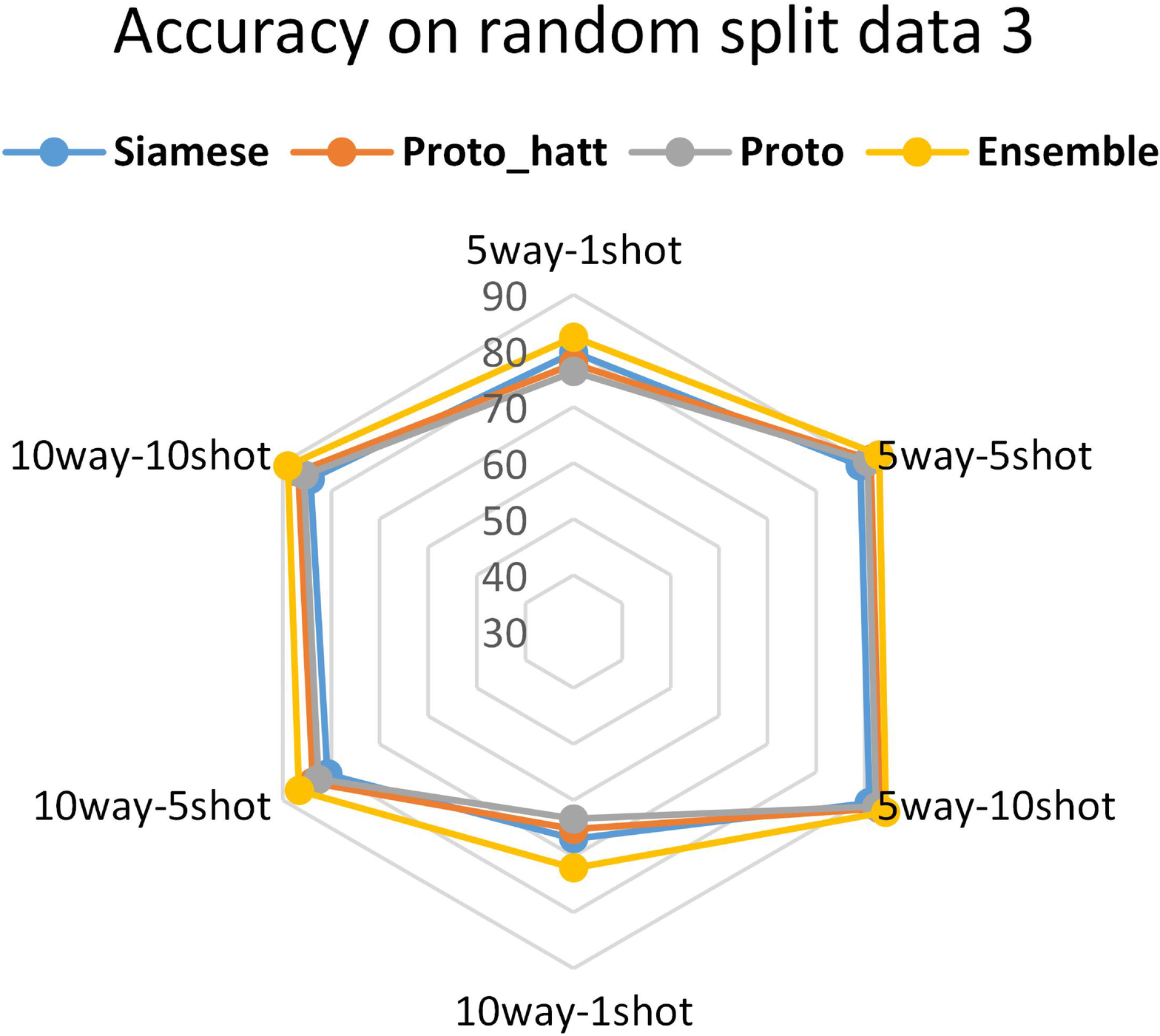}
\end{minipage}
\begin{minipage}[t]{0.5\textwidth}
\centering
\includegraphics[width=0.62\textwidth]{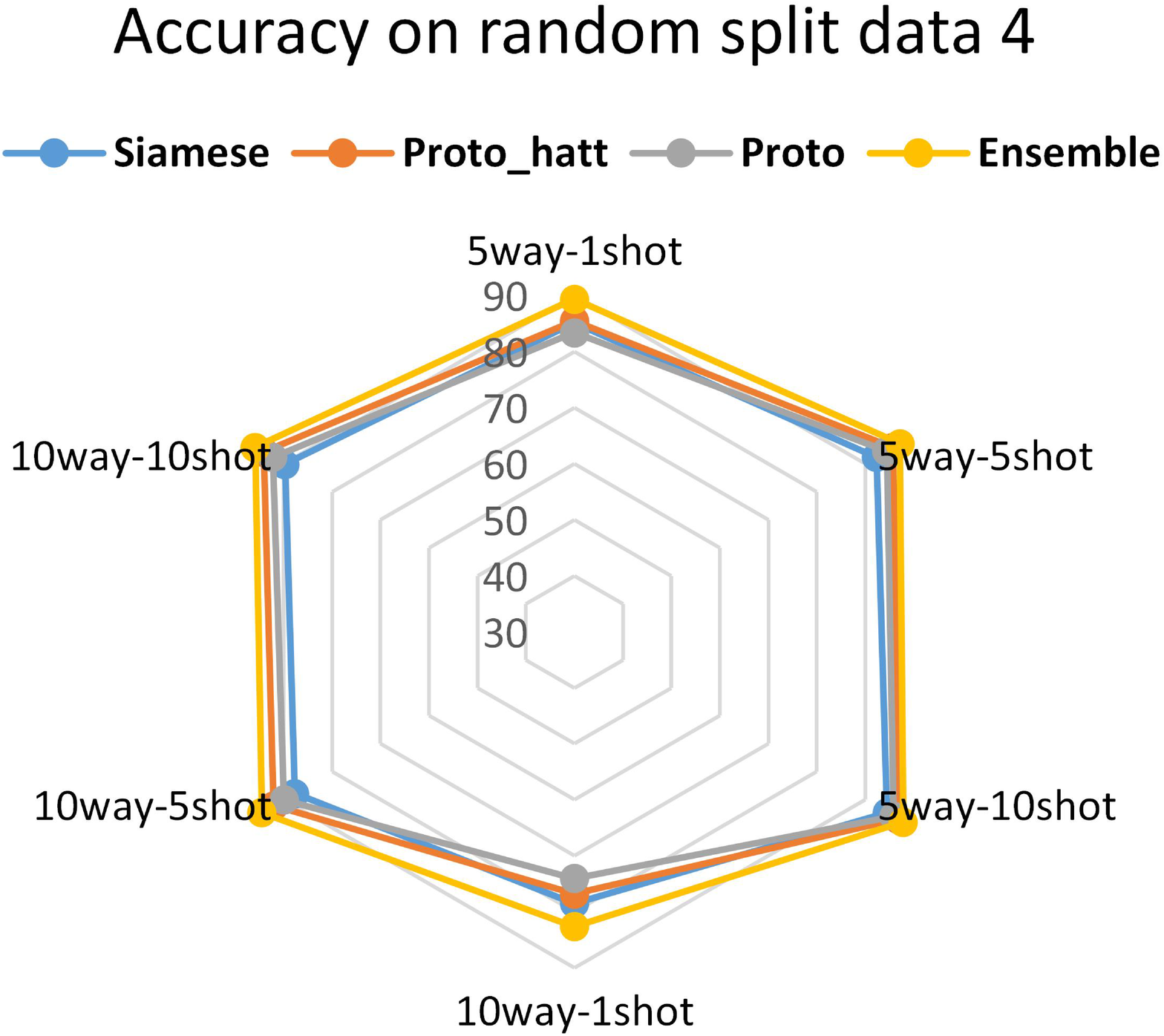}
\end{minipage}
\caption{Radar plots on four Times random split Datasets between the performance of each model and the number of examples in each relation.}
\label{data1-4}
\end{figure*}

In our model, we also use the fine-tuning strategy on k-shot tasks (k$>$1), which is shown on the last line in Table \ref{cross}. The strategy is able to help our model adapt to the tasks in a new domain, and even the result is higher than Bert-pair, which entirely depend on the pre-trained Bert ~\citep{bert}.

Above all, our ensemble model has large improvements on cross domain tasks. Since the few-shot methods on cross domain are still immature, and all of the performances are not good. We perform the other experiments and compare the results on the in-domain tasks. The Table \ref{in} is the overall results on in-domain. The Ensemble\_cosine refers to the results of ensemble on the Cosine metric, the Ensemble\_edis refers to the results of ensemble on the Euclidean metric, and the Ensemble on both.

Focusing on the previous methods in Table \ref{in}, we find that the prototypical network, which is the based architecture of Proto ~\citep{snell2017prototypical03} and Proto\_hatt ~\citep{gao2019hybrid02}, achieves the highest accuracy on 5-shot and 10-shot tasks, only except on 1-shot tasks. However, the performance of the Siamese network is surprised on 1-shot tasks when compared with others. To prove our discovery, we randomly redistribute the relations in our training, validation, and testing set three times and then visualize the experiment results in 
Figure \ref{data1-4}.

As Figure \ref{data1-4} shows that, all of the evaluations on the four times random split dataset have a key similarity: the performance of the Proto model and Proto\_hatt model drop dramatically when the number of examples in each relation is single, which is hard to distinguish the relation label, and the accuracy is lower than other traditional approaches, such as Siamese network which is more stable with the competition results. We analyse the results and find that both the Proto model and Proto\_hatt model depend on the prototype of each relation to predict the relation label. However, in the 1-shot task, the prototype is equal to the only example in each relation; thus, the prototype is completely influenced by noise. Our ensemble model can solve this problem by the cooperation of the component models on 1-shot tasks. On k-shot (k$>=$5) tasks, our model inherits the advantage of the prototypical model, and the other component models in the ensemble model help to recognize more diversity relations. Due to the characteristics demonstrate above, our ensemble model can not only improves the performance on all tasks and becomes more stable when the number of examples changes.

\begin{figure}[h]
\centering
\setlength{\abovecaptionskip}{0.cm}
\setlength{\belowcaptionskip}{-0.cm}
\centering
\includegraphics[width=0.4\textwidth]{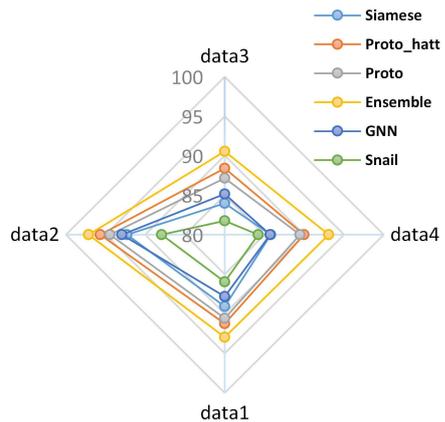}
\caption{Radar plots on 4 times random split datasets to compare the stability of each model}
\label{wz}
\end{figure}

\begin{table*}[h]
    \centering
    \scalebox{0.8}{
    \begin{tabular}{l|lll|lll}
    \hline
         &  & 5 way  &  &  & 10 way  &  \\ 
        Model & 1 shot & 5 shot & 10 shot &  1 shot &  5 shot &  10shot \\ \hline
        Cnn\_edis & 74.31 & 90.02 & 92.32 & 62.03 & 82.35 & 85.88 \\ 
        Cnn\_cosine & 74.63 & 89.08 & 91.16 & 61.63 & 80.90 & 84.08 \\ 
        Incep\_edis & 75.20 & 90.50 & 92.70 & 62.98 & 83.01 & 86.53 \\ 
        incep\_cosine & 77.34 & 90.538 & 92.51 & 65.70 & 83.40 & 86.45 \\ 
        GRU\_edis & 74.12 & 89.83 & 92.01 & 62.42 & 82.56 & 85.83 \\ 
        GRU\_cosine & 76.69 & 90.09 & 91.98 & 65.44 & 82.82 & 85.62 \\ 
        Trans\_edis & 75.29 & 90.13 & 92.14 & 63.73 & 82.79 & 85.73 \\ 
        Trans\_cosine & 75.80 & 89.71 & 91.60 & 64.08 & 82.35 & 85.20 \\ \hline
        Ensemble\_edis & 79.70 & 92.55 & 94.10 & 69.19 & 86.61 & 89.02 \\ 
        Ensemble\_cosine & \textbf{81.40} & 92.56 & 93.98 & 71.22 & 86.72 & 88.90 \\ 
        Ensemble & 81.35 & \textbf{92.90} & \textbf{94.32} & \textbf{71.29} & \textbf{87.30} & \textbf{89.46} \\ \hline
    \end{tabular}}
    \caption{\label{metric}Comparison between metrics with individual encoders}
\end{table*}

To further proves the stability of our ensemble model, we compare the four times experiments horizontally and then map the results of the four experiments to the four dimensions of the radar plots, which is shown in Figure \ref{wz}. The closer distance between each edge and the corresponding equipotential line(grey), the model is more stable and has a lower variance. As we can see, when compared to the other competitive models, our ensemble model achieves the highest accuracy on all of the four experiments. Moreover, we find that the four connected edges of our model are very close to the equipotential lines, which demonstrates that our ensemble model can keep its advantage, ignoring the influence of different relations. In terms of the specific value, the fluctuation ratio of our ensemble model has over 0.5\% decrease, except the Snail network the accuracy of which is disillusionary on all tasks. Moreover, we can find that our ensemble model has more effective on the datasets which is hard to predict. Because the relations are chosen randomly at each time, the above results are sufficient to prove that our method is able to leverage the variance no matter what relations are chosen in training, validation, or testing set. 

Above all, our ensemble model is able to improve the performance, reduce the variance, and enhance domain adaption. Our ensemble approach is necessary and effective for improving robustness.

\subsection{Ablation Study}

In this section, we disassemble our ensemble model to analyze the effectiveness of our approach.

\begin{figure}[h]
\centering
\setlength{\abovecaptionskip}{0.cm}
\setlength{\belowcaptionskip}{-0.cm}
\centering
\includegraphics[width=0.3\textwidth]{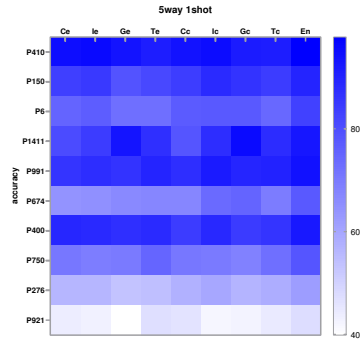}
\caption{Accuracy of each relation in testing set. The deeper color, the higher accuracy.}
\label{hot}
\end{figure}

\textbf{Effect of ensemble different metrics:} In this experiment, we aim show the effect of Euclidean distance and Cosine distance, which are integrated into our ensemble model as the distant metrics between prototypes and query example. In Table \ref{metric}, we compare the performance under the encoders of GRU and Transformers without feature attention. The results demonstrate that the model with cosine distance achieves higher accuracy when the number of examples of each relation is less, while the Euclidean distance is more suitable for the larger number of samples.

To further study, we ensemble the models with Euclidean distance and Cosine
distance, respectively, and the comparison results are presented in Table \ref{metric}. We prove that the above research is also effective in the ensemble model. Besides, the cosine distance is more stable when the number of given examples changes. 

\textbf{Effect of different encoders} In this part, we permutate and combine each encoder and each metrics, and then experiment under 5-way 5-shot setting. We present the results on individual relations as Figure \ref{hot}. We can find that different models are good at predicting different parts of relations. For example, the inception-euclidean model is more suitable to recognize P140, P150, P276, P921 relations when compared with the transformer-cosine model. Thus, each component part of our ensemble model has greatly helped to improve the final performance.

\section{Conclusion}
In this paper, we propose ensemble prototypical
networks for improving accuracy and robustness. Our
ensembles model consist of eight modules, which are basic neural networks. We adopt fine-tuning for enhancing domain adaption and introduce feature attention which alleviates the problem of feature sparsity. In our experiments, we evaluate our model
on FewRel 1.0 and FewRel 2.0, which demonstrate that our model significantly improves the accuracy and robustness, and achieve state-of-the-art results. In the future, we will explore more diverse 
ensemble schemes and adopt more neural encoders to make our model stronger.

\bibliographystyle{acl_natbib}
\bibliography{anthology,acl2021}

\begin{thebibliography}{29}
\expandafter\ifx\csname natexlab\endcsname\relax\def\natexlab#1{#1}\fi

\bibitem[{Chen et~al.(2020)Chen, Wang, Liu, Xu, and Darrell}]{chen2020new}
Yinbo Chen, Xiaolong Wang, Zhuang Liu, Huijuan Xu, and Trevor Darrell. 2020.
\newblock A new meta-baseline for few-shot learning.
\newblock \emph{arXiv preprint arXiv:2003.04390}.

\bibitem[{Cho et~al.(2014)Cho, Van~Merrienboer, Gulcehre, Bahdanau, Bougares,
  Schwenk, and Bengio}]{2014Learning}
Kyunghyun Cho, Bart Van~Merrienboer, Caglar Gulcehre, Dzmitry Bahdanau, Fethi
  Bougares, Holger Schwenk, and Yoshua Bengio. 2014.
\newblock Learning phrase representations using rnn encoder-decoder for
  statistical machine translation.
\newblock \emph{Computer Science}.

\bibitem[{Devlin et~al.(2018{\natexlab{a}})Devlin, Chang, Lee, and
  Toutanova}]{devlin2018bert}
Jacob Devlin, Ming-Wei Chang, Kenton Lee, and Kristina Toutanova.
  2018{\natexlab{a}}.
\newblock Bert: Pre-training of deep bidirectional transformers for language
  understanding.
\newblock \emph{arXiv preprint arXiv:1810.04805}.

\bibitem[{Devlin et~al.(2018{\natexlab{b}})Devlin, Chang, Lee, and
  Toutanova}]{bert}
Jacob Devlin, Ming{-}Wei Chang, Kenton Lee, and Kristina Toutanova.
  2018{\natexlab{b}}.
\newblock {BERT:} pre-training of deep bidirectional transformers for language
  understanding.
\newblock \emph{CoRR}, abs/1810.04805.

\bibitem[{Dhillon et~al.(2020)Dhillon, Chaudhari, Ravichandran, and
  Soatto}]{Dhillon2020ABF}
Guneet~S. Dhillon, P.~Chaudhari, A.~Ravichandran, and Stefano Soatto. 2020.
\newblock A baseline for few-shot image classification.
\newblock \emph{ArXiv}, abs/1909.02729.

\bibitem[{Elsken et~al.(2020)Elsken, Staffler, Metzen, and
  Hutter}]{elsken2020meta}
Thomas Elsken, Benedikt Staffler, Jan~Hendrik Metzen, and Frank Hutter. 2020.
\newblock Meta-learning of neural architectures for few-shot learning.
\newblock In \emph{Proceedings of the IEEE/CVF Conference on Computer Vision
  and Pattern Recognition}, pages 12365--12375.

\bibitem[{Finn et~al.(2017)Finn, Abbeel, and Levine}]{finn2017model18}
Chelsea Finn, Pieter Abbeel, and Sergey Levine. 2017.
\newblock Model-agnostic meta-learning for fast adaptation of deep networks.
\newblock In \emph{Proceedings of the 34th International Conference on Machine
  Learning-Volume 70}, pages 1126--1135. JMLR. org.

\bibitem[{Gao et~al.(2019{\natexlab{a}})Gao, Han, Liu, and
  Sun}]{gao2019hybrid02}
Tianyu Gao, Xu~Han, Zhiyuan Liu, and Maosong Sun. 2019{\natexlab{a}}.
\newblock Hybrid attention-based prototypical networks for noisy few-shot
  relation classification.
\newblock In \emph{Proceedings of the AAAI Conference on Artificial
  Intelligence}, volume~33, pages 6407--6414.

\bibitem[{Gao et~al.(2019{\natexlab{b}})Gao, Han, Zhu, Liu, Li, Sun, and
  Zhou}]{gao2019fewrel09}
Tianyu Gao, Xu~Han, Hao Zhu, Zhiyuan Liu, Peng Li, Maosong Sun, and Jie Zhou.
  2019{\natexlab{b}}.
\newblock Fewrel 2.0: Towards more challenging few-shot relation
  classification.
\newblock In \emph{Proceedings of the 2019 Conference on Empirical Methods in
  Natural Language Processing and the 9th International Joint Conference on
  Natural Language Processing (EMNLP-IJCNLP)}, pages 6251--6256.

\bibitem[{Gao et~al.(2019{\natexlab{c}})Gao, Han, Zhu, Liu, Li, Sun, and
  Zhou}]{gao-etal-2019-fewrel}
Tianyu Gao, Xu~Han, Hao Zhu, Zhiyuan Liu, Peng Li, Maosong Sun, and Jie Zhou.
  2019{\natexlab{c}}.
\newblock \href {https://doi.org/10.18653/v1/D19-1649} {{F}ew{R}el 2.0: Towards
  more challenging few-shot relation classification}.
\newblock In \emph{Proceedings of the 2019 Conference on Empirical Methods in
  Natural Language Processing and the 9th International Joint Conference on
  Natural Language Processing (EMNLP-IJCNLP)}, pages 6250--6255, Hong Kong,
  China. Association for Computational Linguistics.

\bibitem[{Garcia and Bruna(2017)}]{garcia2017few}
Victor Garcia and Joan Bruna. 2017.
\newblock Few-shot learning with graph neural networks.
\newblock \emph{arXiv preprint arXiv:1711.04043}.

\bibitem[{Han et~al.(2018)Han, Zhu, Yu, Wang, Yao, Liu, and
  Sun}]{han2018fewrel08}
Xu~Han, Hao Zhu, Pengfei Yu, Ziyun Wang, Yuan Yao, Zhiyuan Liu, and Maosong
  Sun. 2018.
\newblock Fewrel: A large-scale supervised few-shot relation classification
  dataset with state-of-the-art evaluation.
\newblock In \emph{Proceedings of the 2018 Conference on Empirical Methods in
  Natural Language Processing}, pages 4803--4809.

\bibitem[{Koch et~al.(2015)Koch, Zemel, and Salakhutdinov}]{koch2015siamese14}
Gregory Koch, Richard Zemel, and Ruslan Salakhutdinov. 2015.
\newblock Siamese neural networks for one-shot image recognition.
\newblock In \emph{ICML deep learning workshop}, volume~2. Lille.

\bibitem[{Lecun et~al.(1989)Lecun, Boser, Denker, Henderson, Howard, Hubbard,
  and Jackel}]{2014Backpropagation}
Y~Lecun, B~Boser, J~Denker, D~Henderson, R~Howard, W~Hubbard, and L~Jackel.
  1989.
\newblock Backpropagation applied to handwritten zip code recognition.
\newblock \emph{Neural Computation}, 1(4):541--551.

\bibitem[{Mishra et~al.(2017)Mishra, Rohaninejad, Chen, and
  Abbeel}]{snail2017A}
Nikhil Mishra, Mostafa Rohaninejad, Xi~Chen, and Pieter Abbeel. 2017.
\newblock A simple neural attentive meta-learner.

\bibitem[{Munkhdalai and Yu(2017)}]{munkhdalai2017meta16}
Tsendsuren Munkhdalai and Hong Yu. 2017.
\newblock Meta networks.
\newblock In \emph{Proceedings of the 34th International Conference on Machine
  Learning-Volume 70}, pages 2554--2563. JMLR. org.

\bibitem[{Peters et~al.(2018)Peters, Neumann, Iyyer, Gardner, Clark, Lee, and
  Zettlemoyer}]{Peters2018DeepCW}
Matthew~E. Peters, Mark Neumann, Mohit Iyyer, Matt Gardner, Christopher Clark,
  Kenton Lee, and Luke Zettlemoyer. 2018.
\newblock Deep contextualized word representations.
\newblock In \emph{NAACL-HLT}.

\bibitem[{Qu et~al.(2020)Qu, Gao, Xhonneux, and Tang}]{qu2020few}
Meng Qu, Tianyu Gao, Louis-Pascal Xhonneux, and Jian Tang. 2020.
\newblock Few-shot relation extraction via bayesian meta-learning on relation
  graphs.
\newblock In \emph{International Conference on Machine Learning}, pages
  7867--7876. PMLR.

\bibitem[{Radford(2018)}]{Radford2018ImprovingLU}
A.~Radford. 2018.
\newblock Improving language understanding by generative pre-training.

\bibitem[{Ravi and Larochelle(2017)}]{ravi2016optimization19}
Sachin Ravi and Hugo Larochelle. 2017.
\newblock Optimization as a model for few-shot learning.

\bibitem[{Snell et~al.(2017)Snell, Swersky, and
  Zemel}]{snell2017prototypical03}
Jake Snell, Kevin Swersky, and Richard Zemel. 2017.
\newblock Prototypical networks for few-shot learning.
\newblock In \emph{Advances in neural information processing systems}, pages
  4077--4087.

\bibitem[{Sung et~al.(2018)Sung, Yang, Zhang, Xiang, Torr, and
  Hospedales}]{sung2018learning}
Flood Sung, Yongxin Yang, Li~Zhang, Tao Xiang, Philip~HS Torr, and Timothy~M
  Hospedales. 2018.
\newblock Learning to compare: Relation network for few-shot learning.
\newblock In \emph{Proceedings of the IEEE Conference on Computer Vision and
  Pattern Recognition}, pages 1199--1208.

\bibitem[{Szegedy et~al.(2015)Szegedy, Liu, Jia, Sermanet, Reed, Anguelov,
  Erhan, Vanhoucke, and Rabinovich}]{Szegedy2015GoingDW}
Christian Szegedy, W.~Liu, Y.~Jia, Pierre Sermanet, Scott Reed, Dragomir
  Anguelov, D.~Erhan, V.~Vanhoucke, and Andrew Rabinovich. 2015.
\newblock Going deeper with convolutions.
\newblock \emph{2015 IEEE Conference on Computer Vision and Pattern Recognition
  (CVPR)}, pages 1--9.

\bibitem[{Vanschoren(2018)}]{vanschoren2018meta17}
Joaquin Vanschoren. 2018.
\newblock Meta-learning: A survey.
\newblock \emph{arXiv preprint arXiv:1810.03548}.

\bibitem[{Vaswani et~al.(2017)Vaswani, Shazeer, Parmar, Uszkoreit, Jones,
  Gomez, Kaiser, and Polosukhin}]{vaswani2017attention06}
Ashish Vaswani, Noam Shazeer, Niki Parmar, Jakob Uszkoreit, Llion Jones,
  Aidan~N Gomez, {\L}ukasz Kaiser, and Illia Polosukhin. 2017.
\newblock Attention is all you need.
\newblock In \emph{Advances in neural information processing systems}, pages
  5998--6008.

\bibitem[{Vinyals et~al.(2016)Vinyals, Blundell, Lillicrap, Wierstra
  et~al.}]{vinyals2016matching15}
Oriol Vinyals, Charles Blundell, Timothy Lillicrap, Daan Wierstra, et~al. 2016.
\newblock Matching networks for one shot learning.
\newblock In \emph{Advances in neural information processing systems}, pages
  3630--3638.

\bibitem[{Wei-Yu et~al.(2019)Wei-Yu, Liu, Kira, Wang, and
  Huang}]{chen2019closerfewshot}
Chen Wei-Yu, Yen-Cheng Liu, Zsolt Kira, Yu-Chiang Wang, and Jia-Bin Huang.
  2019.
\newblock A closer look at few-shot classification.
\newblock In \emph{International Conference on Learning Representations}.

\bibitem[{Ye and Ling(2019)}]{ye2019multi}
Zhi-Xiu Ye and Zhen-Hua Ling. 2019.
\newblock Multi-level matching and aggregation network for few-shot relation
  classification.
\newblock \emph{arXiv preprint arXiv:1906.06678}.

\bibitem[{Yoon et~al.(2018)Yoon, Kim, Dia, Kim, Bengio, and
  Ahn}]{yoon2018bayesian}
Jaesik Yoon, Taesup Kim, Ousmane Dia, Sungwoong Kim, Yoshua Bengio, and Sungjin
  Ahn. 2018.
\newblock Bayesian model-agnostic meta-learning.
\newblock \emph{Advances in Neural Information Processing Systems},
  31:7332--7342.

\end{thebibliography}


\newpage

\section*{Appendices}

\subsection{Open Relation Extraction Datasets}

Download: https://github.com/thunlp/FewRel

\subsection{Computing Infrastructure}

Computing infrastructure:  GPU Tesla V100

\subsection{Encoders Description}

\textbf{Encoder module}

Given an sample $x$, represented by a sequence of word embeddings, we use different neural architectures as sentence encoders to get a continuous low-dimensional sample embedding $\mathrm{x}\epsilon \mathbb{R}^{D}$. We denote the encoder operation as the following equation.

\begin{equation}
\mathrm{x}=f_{\theta}(x)
\end{equation}

Where $\theta$ represents the neural network architectures used by encoder operation.

\textbf{CNN encoder}: In this encoder module, we select CNN to encode $x$ into an embedding . In the CNN, convolution and pooling operation are successively applied to capture the text semantics and get sample embedding

\begin{equation}
\begin{array}{c}
H=\operatorname{conv}(w,x)+b \\
\mathrm{x}=\operatorname{pool}(H)
\end{array}
\end{equation}

Where convolution operation $conv(.)$ uses a convolution kernel $w$ to slide over the word embedding to get hidden embeddings, and pooling operation $pool(.)$ uses Max pooling to output the final samples embedding $\mathrm{x}$. We simplify the above operation to the following equation:

\begin{equation}
\mathrm{x}=f_{cnn}(x)
\end{equation}

\textbf{Inception encoder:} Referring to the GoogLeNet, we design a inception module with wider convolution layer as the encoder, which uses multiple parallel convolution kernels with different window size $w(1*1,3*3,5*5,7*7,9*9)$ to encode sample $x$ to get hidden embeddings.

\begin{equation}
\begin{array}{l}
H_{1}=\operatorname{conv}\left(w_{1 * 1}, x\right)+b \\
H_{2}=\operatorname{conv}\left(w_{3 * 3}, x\right)+b \\
H_{3}=\operatorname{conv}\left(w_{5 * 5}, x\right)+b \\
H_{4}=\operatorname{conv}\left(w_{7 *7}, x\right)+b \\
H_{5}=\operatorname{conv}\left(w_{3 * 3} \operatorname{conv}\left(w_{3+3}, x\right)+b\right)+b
\end{array}
\end{equation}

Where $w_{i*i}$ means to use a convolution kernel with the window size i for convolution operation. In order to reduce the computational complexity of convolution operation with $9*9$ size convolution kernel, we decompose it into two $3*3$ size convolution kernels.

The features obtained from different scale convolution operations are fused to get the final hidden embeddings

\begin{equation}
\mathrm{H}=\operatorname{concat}\left(H_{1}, H_{2}, H_{3}, H_{4}, H_{5}\right)
\end{equation}

Here $concat(.)$ means to concatenate all embeddings ($H_{1},., H_{5}$) into a higher-dimensional embeddings $\mathrm{H}$.

Finally, we get the sample embedding $\mathrm{x}$ by applying a pooling operation on hidden embeddings $\mathrm{H}$

\begin{equation}
\mathrm{x}=\operatorname{pool}(\mathrm{H})
\end{equation}

We demote the above operation as the following equation:

\begin{equation}
\mathrm{x}=f_{\text {incep }}(x)
\end{equation}

\textbf{Base-attention GRU encoder:} in this encoder, we use a bi-direction recurrent neural network with self-attention to process the sample $x$. The encoder consists of a bidirectional GRU layer and a self-attention layer.

GRU layer uses a parameter-shared gru cell to process the samples $x(e_{1},...,e_{n})$ to get the hidden embeddings $H(h_{1},...,h_{n})$ based on the current input state $e_{t}$ and the previous output state $h_{t-1}$

\begin{equation}
\begin{array}{l}
\overrightarrow{h_{n}}=GRU\left(e_{n}, h_{n-1}\right) \\
\overleftarrow{h_{n}}=GRU\left(e_{n}, h_{n-1}\right) \\
h_{n}=<\overrightarrow{h_{n}},\overleftarrow{h_{n}}>
\end{array}
\end{equation}

we concatenate $\overrightarrow{h_{n}}$ and $\overleftarrow{h_{n}}$ to get hidden embeddings $h_{n}$ and denote all h as the $H(h_{1},...,h_{n})$, and then we obtain final sample embedding $\mathrm{x}$ by linear combination of hidden embeddings in $\mathrm{H}$. The self-attention layer is used to compute the linear combination, which takes the hidden embeddings in $\mathrm{H}$ as input and computes the weight vector a

\begin{equation}
\begin{array}{c}
\mathrm{a}=\operatorname{softmax}\left(\mathrm{t}(\mathrm{g}(\mathrm{H}))\cdot \mathrm{t}(\mathrm{g}(\mathrm{H}))\right) \\
\mathrm{A}=a^{T} H
\end{array}
\end{equation}

Where $g(.)$ is a linear layer, $t(.)$ is an activation function. the final representation $\mathrm{x}$ of samples is the sum of $A(a_{1},...,a_{n})$

\begin{equation}
\mathrm{x}=\sum_{n=1}^{n} a_{n}
\end{equation}

We demote the above operation as the following equation:

\begin{equation}
\mathrm{x}=f_{GRU}(x)
\end{equation}

\textbf{Base-attention Transformer encoder:} Transformer is a parallel computing neural network structure composed of attention mechanism, which consists of two part: Encoder and Decoder. In this encoder, we use Transformer Encoder as our sentence encoder.

\begin{equation}
H=\operatorname{trans}(x)
\end{equation}

Transformer encoder takes the word embeddings of samples x as the input, and output the hidden embeddings $H$. Then we append a self-attention layer like in GRU to compute the linear combination of hidden embedding $h_{i}$ in H and get the final sample embedding $\mathrm{x}$.

\begin{equation}
\mathrm{x}=\operatorname{Atten}(H)
\end{equation}

We demote the above operation as the following equation:

\begin{equation}
\mathrm{x}=f_{trans}(x)
\end{equation}

\subsection{Other Experimental Results}

Experiment result on the cross-domain dataset Fewrel 2.0

\begin{table*}
    \centering
    \begin{tabular}{l|lll|lll}
    \hline
         &  & 5 way  &  &  & 10 way  &  \\
        Model & 1 shot & 5 shot & 10 shot &  1 shot &  5 shot &  10shot \\ \hline
        Cnn\_edis & 41.54 & 55.87 & 62.28 & 29.69 & 42.33 & 48.62 \\
        Cnn\_cosine & 40.57 & 51.95 & 57.45 & 28.56 & 38.13 & 43.05 \\
        Incep\_edis & 39.70 & 55.11 & 61.81 & 27.82 & 41.57 & 48.10 \\
        incep\_cosine & 39.70 & 53.69 & 60.07 & 27.93 & 40.14 & 46.16 \\
        GRU\_edis & 40.41 & 53.73 & 58.99 & 28.28 & 40.07 & 45.05 \\
        GRU\_cosine & 39.63 & 52.55 & 57.68 & 27.79 & 39.01 & 43.99 \\
        Trans\_edis & 41.15 & 54.98 & 60.43 & 29.14 & 41.40 & 46.89 \\
        Trans\_cosine & 40.16 & 53.61 & 59.22 & 28.24 & 40.19 & 45.65 \\
        Ensemble & \textbf{44.42} & \textbf{61.76} & \textbf{68.49} & \textbf{32.44} & \textbf{48.26} & \textbf{55.23} \\ \hline
    \end{tabular}
    \caption{the accuracy of each sub-model in our ensemble method on the cross-domain dataset Fewrel2.0}
\end{table*}

\begin{table*}
    \centering
    \begin{tabular}{l|lll|lll}
    \hline
         &  & 5 way  &  &  & 10 way  &  \\
        Model & 1 shot & 5 shot & 10 shot &  1 shot &  5 shot &  10shot \\ \hline
        Ensemble\_cosine & 42.67 & 57.99 & 64.45 & 30.55 & 44.06 & 50.43 \\
        Ensemble\_edis & 43.24 & 59.70 & 66.20 & 31.13 & 45.95 & 52.58 \\
        Ensemble & \textbf{44.39} & \textbf{61.27} & \textbf{67.89} & \textbf{32.20} & \textbf{47.61} & \textbf{54.41} \\ \hline
    \end{tabular}
    \caption{The result of our ensemble model that uses the voting method as the ensemble scheme on the cross-domain dataset}
\end{table*}

\begin{table*}
    \centering
    \begin{tabular}{l|lll|lll}
    \hline
         &  & 5 way  &  &  & 10 way  &  \\
        Model & 1 shot & 5 shot & 10 shot &  1 shot &  5 shot &  10shot \\ \hline
        Cnn\_edis & / & 58.55 & 66.28 & / & 48.01 & 56.05 \\
        Cnn\_cosine & / & 61.28 & 68.94 & / & 53.17 & 62.94 \\
        Incep\_edis & / & 63.10 & 70.26 & / & 57.05 & 65.52 \\
        incep\_cosine & / & 60.02 & 67.91 & / & 51.36 & 60.14 \\
        GRU\_edis & / & 57.91 & 63.03 & / & 49.53 & 56.32 \\
        GRU\_cosine & / & 56.43 & 61.90 & / & 47.97 & 54.36 \\
        Trans\_edis & / & 63.35 & 69.88 & / & 57.93 & 63.91 \\
        Trans\_cosine & / & 62.93 & 68.49 & / & 57.85 & 65.15 \\
        Ensemble\_fine-tune & / & \textbf{68.32} & \textbf{74.96} & / & \textbf{63.98} & \textbf{70.61} \\ \hline
    \end{tabular}
    \caption{the accuracy of each sub-model after fine-tuning on the cross-domain dataset Fewrel2.0}
\end{table*}

Comparative experiment in which the model uses Glove word vector and Bert word vector as initial word embedding respectively on the random split in-domain  FewRel 1.0. The accuracy of the model fluctuates greatly when Bert word vector was used. Here, So, is only a preliminary comparison.

\begin{table*}
    \centering
    \begin{tabular}{l|ll}
    \hline
        Model & GloVe & Bert \\ \hline
        Cnn\_edis & 89.81 & 92.78 \\
        Cnn\_cosine & 89.28 & 91.93 \\
        Incep\_edis & 91.01 & 93.65 \\
        incep\_cosine & 90.62 & 92.75 \\
        GRU\_edis & 90.49 & 91.84 \\
        GRU\_cosine & 89.79 & 90.63 \\
        Trans\_edis & 90.93 & 92.06 \\
        Trans\_cosine & 90.62 & 91.712 \\
        Ensemble & 93.18 & \textbf{94.71} \\
        Bert-pair & 94.19 & \\ \hline
    \end{tabular}
    \caption{Comparison between our models that used Glove word vector and Bert word vector as initial word embedding respectively}
\end{table*}

\end{document}